%% file: main.tex
\documentclass[10pt,twocolumn,letterpaper]{article}

\usepackage[pagenumbers]{cvpr} 

\input{preamble}
\input{ALL_FIGURES}

\definecolor{cvprblue}{rgb}{0.21,0.49,0.74}
\usepackage[pagebackref,breaklinks,colorlinks,allcolors=cvprblue]{hyperref}

\def\confName{XXXX}
\def\confYear{XXXX}

\title{What's in the box? Revealing Object Layouts Inside Opaque Boxes via Contactless Vibration Excitation and Sensing}

\author{Ethan Turok \qquad Matan Kichler \qquad Shai Bagon \qquad Mark Sheinin\\
Weizmann Institute of Science, Israel\\
{\tt\small \{ethan.turok,matan.kichler,shai.bagon,mark.sheinin\}@weizmann.ac.il}\\
}

\begin{document}
\maketitle
\input{sec/0_abstract}    

\input{sec/1_intro}
\input{sec/2_related_works}
\input{sec/3_method}
\input{sec/4_details}
\input{sec/5_experimental}

\input{sec/6_insights}
\input{sec/7_discussion}
\input{sec/8_conclude}

\iftoggle{cvprfinal}{
\paragraph{Acknowledgments}
This project received funding from the NSF-BSF grant \#2024808, 
the MBZUAI-WIS Joint Program for AI Research, 
and the Knell Family Institute for Artificial Intelligence (SB).
}{}

{
     \small
     \bibliographystyle{ieeenat_fullname}
     \bibliography{main}
}

\end{document}

%% file: preamble.tex
\definecolor{cvprblue}{rgb}{0.21,0.49,0.74}
\usepackage[pagebackref,breaklinks,colorlinks,allcolors=cvprblue]{hyperref}

\usepackage{siunitx}
\usepackage{bm}
\usepackage{graphicx}
\usepackage{amsmath}
\usepackage{amssymb}
\usepackage{booktabs}
\usepackage{comment}

\usepackage{url}
\usepackage{svg}
\usepackage[table]{xcolor}
\usepackage{xspace}
\usepackage{ifthen}
\usepackage[normalem]{ulem}

\graphicspath{{figures_overleaf/}}
\usepackage{xspace}
\usepackage{ifthen}
\usepackage[normalem]{ulem}

\newcommand{\MS}[2][]{%
    \ifthenelse{ \equal{#1}{} }
        {\textcolor{red}{[MS] #2\xspace{}}}
        {\textcolor{red}{[MS] \sout{#1} #2\xspace{}}}
}

\definecolor{darkred}{rgb}{0.75, 0.1, 0.5}
\newcommand{\ET}[2][]{%
    \ifthenelse{ \equal{#1}{} }
        {\textcolor{darkred}{[ET] #2\xspace{}}}
        {\textcolor{darkred}{[ET] \sout{#1} {#2}\xspace{}}}
}

\newcommand{\SB}[2][]{%
    \ifthenelse{ \equal{#1}{} }
        {\textcolor{orange}{#2\xspace{}}}
        {\textcolor{orange}{\sout{#1} #2\xspace{}}}
}

\def\capfont{\normalfont\small}

\newcommand{\afterfigure}{\vspace*{-10pt}}

%% file: ALL_FIGURES.tex
\newcommand{\figTeaser}{
    \begin{figure}[t]
    \centering
    \includegraphics[width=\linewidth]{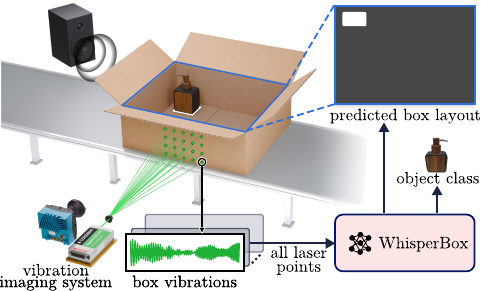}
    \caption{Seeing inside opaque boxes with contactless vibrations. Our system uses a speaker to excite the box and  a remote vibration imaging system to capture its vibrations. We train WhisperBox, a deep learning model that reconstructs the hidden box layout from these vibrations without touching or moving the box.}
    \afterfigure
    \label{fig:teaser}
    \end{figure}
}

\newcommand{\figExperimentSetup}{
    \begin{figure}[t]
    \centering
        \includegraphics[width=\linewidth]{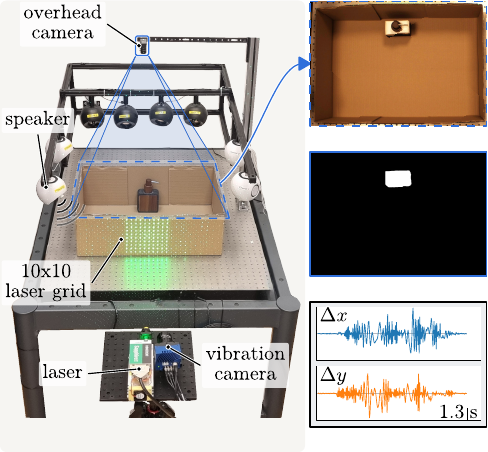}
        \put(-205, 3){\capfont \textbf{(a)} experimental setup}
        \put(-86, 149){\capfont \textbf{(b)} box interior image}
        \put(-86, 76){\capfont \textbf{(c)} ground truth layout}
        \put(-86, 3){\capfont \textbf{(d)} raw point vibrations}
    \caption{Our data collection setup.
        \textbf{(a)} A speaker excites the box while a vibration camera records the motion of a $10 \times 10$ laser grid projected onto its wall. An overhead camera captures
        \textbf{(b)} the box interior, from which we extract
        \textbf{(c)} the binary box layout.
        \textbf{(d)}  \textbf{(d)} Vibration of one laser point, as $x$ and $y$ pixel shifts over $1.3$s audio.
        In our dataset, each sample contains 100 laser point vibrations like \textbf{(d)} (model input) and the box layout \textbf{(c)} (model output).
        }
    \vspace{-2pt}
    \afterfigure
    \label{fig:setup}
    \end{figure}
}

\newcommand{\figNetworkArch}{
\begin{figure*}[!tp]
    \centering
    \includegraphics[width=1\textwidth]{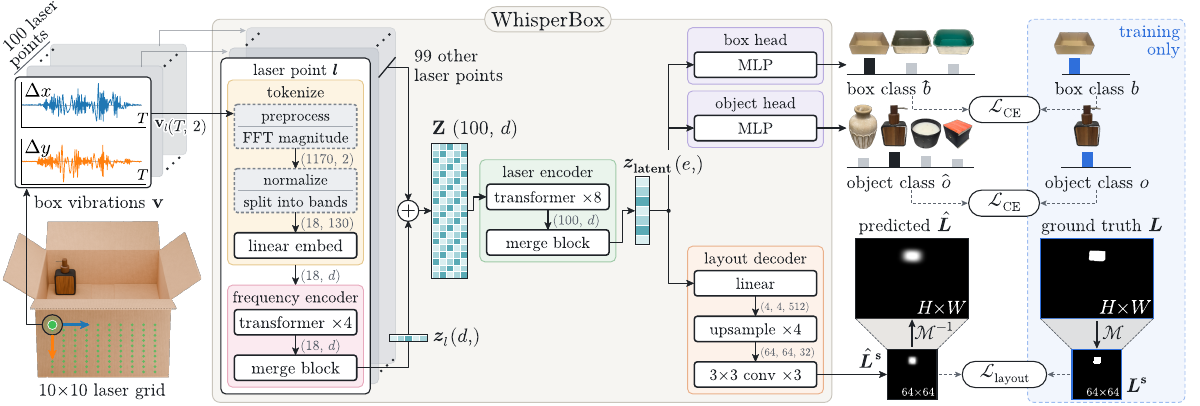}
    \caption{\textbf{WhisperBox network architecture}. Our model takes vibration measurements from $100$ laser points as input. The tokenizer converts each measurement into $18$ frequency tokens. The frequency encoder processes each laser point's tokens independently, learning frequency-dependent features. The laser encoder then combines information across all 100 points into a shared representation of the hidden scene. A decoder reconstructs the 2D occupancy layout $\hat{L}^s$, while auxiliary heads predict the box $\hat{b}$ and object $\hat{o}$ identities. Solid white boxes denote learned weights, and gray dashed boxes denote fixed transformations. We use $d=256$ and $e=512$.}
    \afterfigure
    \label{fig:network_arch}
\end{figure*}
}

\newcommand{\figQualitativeResults}{
    \begin{figure*}[t]
    \centering
    \includegraphics[width=\linewidth]{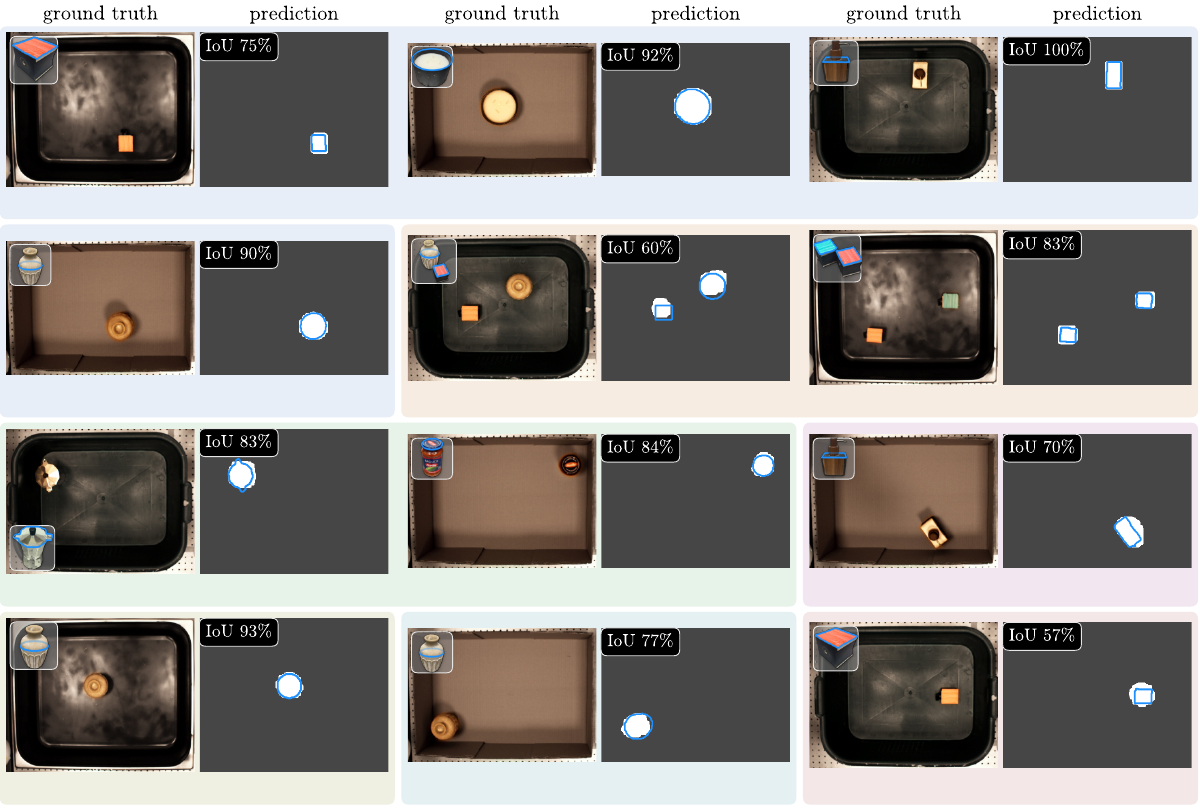}%
    \put(-494.2, 247.4){\parbox[t]{158.4pt}{\centering\capfont \textbf{(a) in distribution:} cube in metal}}%
    \put(-327.6, 247.4){\parbox[t]{158.4pt}{\centering\capfont \textbf{in distribution:} candle in cardboard}}%
    \put(-161.1, 247.4){\parbox[t]{158.4pt}{\centering\capfont \textbf{in distribution:} soap in plastic}}%
    \put(-494.2, 165.3){\parbox[t]{158.4pt}{\centering\capfont \textbf{in distribution:} vase in cardboard}}%
    \put(-327.6, 165.3){\parbox[t]{158.4pt}{\centering\capfont \textbf{(b) two objects:} cube and vase in plastic}}%
    \put(-161.1, 165.3){\parbox[t]{158.4pt}{\centering\capfont \textbf{two objects:} two cubes in metal}}%
    \put(-494.2, 86.8){\parbox[t]{158.4pt}{\centering\capfont \textbf{(c) unseen object:} coffee pot in plastic}}%
    \put(-327.6, 86.8){\parbox[t]{158.4pt}{\centering\capfont \textbf{unseen object:} tomato sauce in cardboard}}%
    \put(-161.1, 86.8){\parbox[t]{158.4pt}{\centering\capfont \textbf{(d) unseen rotation:} soap in cardboard}}%
    \put(-494.2, 4.7){\parbox[t]{158.4pt}{\centering\capfont \textbf{(e) unseen box region:} vase in metal}}%
    \put(-327.6, 4.7){\parbox[t]{158.4pt}{\centering\capfont \textbf{(f) unseen speaker:} vase in cardboard}}%
    \put(-161.1, 4.7){\parbox[t]{158.4pt}{\centering\capfont \textbf{(g) unseen laser speckles:} cube in plastic}}%
    \caption{
    Qualitative WhisperBox results.
    Each pair shows the overhead ground-truth photo (left) and the predicted layout (right). Blue outlines mark the object's contour from the overhead photo.
    \textbf{(a) In distribution:} objects seen in training, at held-out positions.
    \textbf{(b) Two objects:} two objects at once; these pairs are seen in training, at held-out positions.
    \textbf{(c) Unseen object:} objects never seen in training.
    \textbf{(d) Unseen rotation:} a training object at a rotated orientation held out from training.
    \textbf{(e) Unseen box region:} a training object placed in a part of the box where it never appears in training.
    \textbf{(f) Unseen speaker:} a training object recorded with a speaker held out from training.
    \textbf{(g) Unseen laser speckles:} a training object recorded with a shifted laser grid: the lasers hit points on the box wall not measured in training.
    }
    \label{fig:good_predictions}
    \vspace{-5pt}
    \afterfigure
    \end{figure*}
}

\newcommand{\figQuantitativeResults}{
    \begin{figure}[t]
        \centering
        \includegraphics[width=\columnwidth]{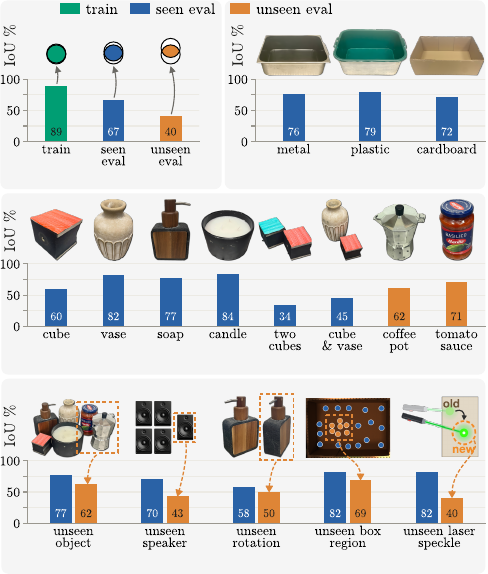}
        \put(-230, 190){\capfont \textbf{(a)} overall performance}
        \put(-119, 190){\capfont \textbf{(b)} mean performance by box}
        \put(-170, 99){\capfont \textbf{(c)} mean performance by object }
        \put(-190, 3){\capfont \textbf{(d)} mean out-of-distribution performance}
    \caption{
   \textbf{Quantitative WhisperBox results by IoU}.
    \textbf{(a)} WhisperBox generalizes well to seen settings and moderately to unseen ones.
    \textbf{(b)} Performance is consistent across all boxes.
    \textbf{(c)} Performance varies by object: two objects scenes are the hardest followed by unseen objects.
    \textbf{(d)} Among unseen settings, WhisperBox generalizes well to new objects, rotations, and box regions, but less so to new speakers and laser speckles.
    }
    \vspace{-5pt}
    \afterfigure
    \label{fig:plots}
    \end{figure}
}

\newcommand{\figInsights}[1]{
    \begin{figure}[t]
        \centering
        \includegraphics[width=\columnwidth]{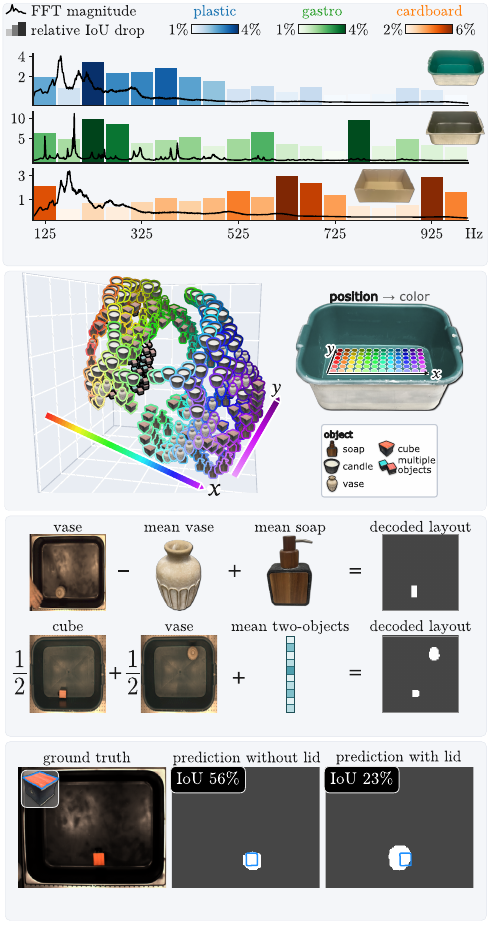}
        \put(-205, 326){\capfont \textbf{(a)} spectral vibration magnitude versus importance}
        \put(-170, 205){\capfont \textbf{(b)} latent space structure}
        \put(-170, 97){\capfont \textbf{(c)} latent space arithmetic}
        \put(-200, 9){\capfont \textbf{(d)} prediction with and without a lid on the box}
    \vspace{-5pt}
    \caption{Results analysis.
    \textbf{(a)} Plot showing that the frequency band most contributing to correct layout predictions does not necessarily coincide with the boxes' resonance frequencies.
    \textbf{(b)} Latent space shows structure related to object location within the box.
    \textbf{(c)} The latent space supports arithmetic operations.
    \textbf{(d)} Closing the box lid worsens the prediction.
    }
    \vspace{-8pt}
    \afterfigure
    \label{fig:analysis}
    \end{figure}
}

%% file: sec/0_abstract.tex
\begin{abstract}
Can we reveal what is inside an opaque box by observing its surface vibrations? Objects inside a box alter its mechanical response, providing cues about their hidden locations. We introduce a method that uses these cues to predict a box's 2D occupancy layout entirely remotely. A nearby speaker excites vibrations, which we measure on one visible wall using a grid of laser points and a camera measuring each point's speckle-pattern shifts. Our model, WhisperBox, learns to encode these vibrations into a latent representation and decode it into an occupancy map of the box's contents. The method requires no physical contact to excite or sense vibrations. We demonstrate layout recovery across various boxes and single- and multiple-object scenes, with generalization to unseen object instances, positions, orientations, and weights, as well as changes in laser placement and excitation source. Our analysis reveals that informative frequency bands need not coincide with the box's resonances and that the learned latent space captures object placement and supports arithmetic operations. These results demonstrate the potential of remote vibration sensing to infer spatial layouts hidden from view.
\end{abstract}

%% file: sec/1_intro.tex
\section{Introduction}
\label{sec:intro}

Computer vision derives immense utility from its ability to remotely detect, classify, and localize objects in a scene. However, predominantly, this ability is limited to objects within the line of sight of conventional RGB cameras. In this work, we explore the exciting prospect of remotely detecting and localizing objects within closed containers and boxes that are completely opaque to visible light (\eg, a plastic container and a cardboard box). 
This ability is highly desired for security (\eg,~detecting threats in luggage), industrial inspection (\eg,~detecting missing or broken items in packaging) and robotic manipulation (\eg,~locating concealed objects before opening a container). But how can we overcome the opaqueness of the observed container? 

One obvious solution is to use wavelengths that \textit{do} penetrate the occluding materials.
X-rays are used to detect threats in luggage \cite{wei2020occluded, ren2025prohibited}, contaminants sealed in food packaging \cite{kim2021real}, and defects hidden in assembled electronics \cite{wankerl2020fully}. However, X-rays' ionizing radiation is dangerous to humans, requiring fully shielded enclosures that limit where they can be used. 
Radio-frequency (RF) signals are safer and also pass through walls and boxes. For instance, Wifi signals were used to predict the 3D pose of humans behind a wall \cite{adib2013see, wang2019person, zhao2018through} and millimeter-wave signals are used to classify objects inside envelopes \cite{ganse2021high} and generate 2D images and 3D point clouds of objects hidden inside boxes \cite{doddsmito, dodds2026wave, lu2026geraf}. However, mm-waves are noisy \cite{dodds2026wave} and specular (mirror-like), reflecting most of the signal away from the sensor. This leaves parts of the object unobserved and requires custom shape-completion methods \cite{dodds2026wave}. Moreover, neither X-ray nor Wi-Fi signals can penetrate metal boxes.

\figTeaser

While visible-light imaging cannot penetrate a container, it \textit{can} measure a proxy signal on the container's surface that can reveal its contents: the container's vibrations. 
Several seminal works used high-speed cameras to infer mechanical properties from visible vibrations \cite{bouman2013estimating,davis2015visual} and recover sound from vibrating objects \cite{davis2014visual}. Later works extended this approach to estimate internal material properties and thickness \cite{feng2022visual,ogren2025visual}.
However, high-speed cameras are expensive, bandwidth-limited, and struggle to capture low-amplitude, high-frequency vibrations. Recently, active, laser-based methods have enabled reliable, high-speed capture of surface vibrations at multiple object points at once \cite{sheinin2022dual,kichler2025learning}. Researchers have used these visual vibrometry cameras to estimate scene audio \cite{bagon2026hearing,sheinin2022dual,cai2025event2audio}, localize impacts \cite{zhang2023analyzing}, and visualize dense surface vibrations \cite{chen2024coherence}. Kichler \etal first employed such active sensing to infer the fill level of opaque liquid containers \cite{kichler2025learning}. Our work extends this semantic, vibration-based content inference from a 1D quantity (\ie,~fill level) to a far more complex and challenging spatial representation: a 2D occupancy layout.

In this paper, we develop a learning-based approach to infer a container's 2D occupancy layout from vibrations measured on one visible side. Specifically, we place various objects within the container and measure its response to vibrations excited by a nearby speaker on a 10x10 grid of laser points. Then, we input these raw vibrations into our model, which learns to encode the high-dimensional temporal signals into a single latent embedding vector. Subsequently, a convolution-based decoder learns to predict the container's occupancy map from the latent representation. The model is agnostic to the box's dimensions and scale.

During training, we use an overhead camera to capture the container's interior (not visible to our vibration camera) and produce a segmentation mask of the container's true layout. 
We gather a dataset of various containers and objects and train our model to predict the container's layout for new object positions solely from the captured vibrations.

We experimentally validate our approach by predicting unseen object positions with high fidelity for the objects in our training set. Additionally, we show that the model generalizes beyond the training domain by predicting unseen object instances, orientations, weights, and box regions. Moreover, we show that the model can predict a box's layout when the laser strikes a different region of the box wall. Our model can also recover scenes with multiple objects in the same box and use new unseen speakers.

Beyond experimental performance, we provide an in-depth analysis of our model's learned latent space. Specifically, we show that our model's predictions rely on frequency bands that do not coincide with the box's resonance frequencies. We visualize the structure of our model's embedding space, showing it learned to map the object's geometric placement despite only being trained to predict layouts. Lastly, we show that the learned latent space supports arithmetic operations, similar to those of standard text embeddings. 

%% file: sec/2_related_works.tex
\section{Related works}
\label{sec:related}

\begingroup
\setlength{\parskip}{4pt}

\noindent\textbf{Optical non-line-of-sight imaging.}
Optical non-line-of-sight (NLOS) imaging recovers hidden scenes from light indirectly reaching a camera through visible relay surfaces. The first method iterations demonstrated reconstruction and tracking around corners \cite{kirmani2009looking,velten2012recovering,heide2014diffuse,gariepy2016detection}. Subsequent advances introduced efficient light-cone and wave-based reconstructions \cite{otoole2018confocal,lindell2019wave,liu2019phasor}, temporal focusing \cite{pediredla2019snlos}, and methods that exploit surface orientation, occlusions, and light-path geometry \cite{heide2019partial,xin2019fermat,tsai2019beyond,young2020directional}. Some approaches recover hidden information from intensity measurements using conventional cameras \cite{bouman2017turning,saunders2019periscopy,chen2019steady,sharma2021blank}. Neural scene representations \cite{shen2021neural} and imaging through a single optical access point \cite{metzler2021keyhole} further broaden these capabilities. Like these methods, our approach infers hidden spatial structure by observing a visible surface. However, optical NLOS methods require a light path connecting the hidden scene to the sensor, whereas we measure the enclosure's mechanical response, enabling inference inside sealed, optically opaque containers without optical access to their contents.

\noindent\textbf{Inference from airborne sound.}
Standard microphones have been used to determine the amount of liquid in a bottle from the sound of it being poured \cite{liang2019making, wilson2019analyzing, bagad2026sound}, infer an object's shape and material from the sound of it falling or being struck \cite{zhang2017shape, clarke2022diffimpact}, and even recover depth maps and grayscale images of a room from its echoes \cite{christensen2020batvision, gao2020visualechoes}. However, conventional microphones aggregate sounds from across the environment, struggle to capture faint sounds from distant objects because acoustic intensity decreases with distance, and cannot directly measure vibrations at a specific point on an object.

\noindent\textbf{Contact-based inference of hidden contents.}
In contrast, contact microphones can measure vibrations directly on some object surface. Prior work has used contact microphones to identify different objects \cite{liu2017vibsense}, distinguish between various touch events (tap, swipe, etc.) \cite{iravantchi2023sawsense}, and even reconstruct terrain from footstep vibrations \cite{liu2025wildfusion}. Recently, researchers have placed contact microphones on robotic hands, enabling object material classification \cite{liu2024sonicsense, lu2023active}, contact localization \cite{lu2023active, zai2026vibro, yi2409visual, lee2025sonicboom}, and even 3D object shape reconstruction \cite{liu2024sonicsense}.

Contact microphones have also been used in containers to identify and track objects sliding inside boxes \cite{gandhi2020swoosh}, estimate granular material in a vibrating scooper \cite{clarke2018learning}, and distinguish dice counts or water amounts in shaken containers \cite{liu2024sonicsense}. Closest to our work, Boombox \cite{chen2021boombox} uses contact microphones to reconstruct an RGB-D image of a box’s contents from vibrations caused by an object dropped inside. However, it requires both physical contact for sensing and perturbation of the container to generate a signal. 
Our work, named WhisperBox in tribute to Boombox \cite{chen2021boombox}, removes both requirements and recovers the contents of a container without touching it, either to excite vibrations or to sense them. This enables \textit{fully remote sensing} of occluded objects in sealed containers as they are: nothing inside is dropped or moved, and no sensor is attached to the container.
\par
\endgroup

%% file: sec/3_method.tex
\section{Revealing Object Layouts using Vibration}
\label{sec:method}

\paragraph{Problem overview and scope.}
We seek to predict the content of a box from its vibrations alone. This task relies on the fact that the box's vibrations change, ever so slightly, when an external mass presses on an area of the box's floor. This is similar to how resting a finger on a drumhead changes its vibrations (and resulting sound). Recovering the internal layout from these vibrations requires solving a coupled inverse problem involving the box's geometry, mass distribution, stiffness, boundary conditions, object--container contacts, and spatially varying acoustic excitation. These factors are generally unknown and difficult to calibrate for real everyday containers, making direct physics-based inversion impractical. We therefore develop a learning-based approach to infer the internal layout from the measured vibrations.

At inference time, we excite the box using a nearby external speaker playing a chirp. Our vibration camera measures the vibrations at a grid of points of the box's side. During training, an additional overhead camera records the ground-truth box layout.
We assume that the box's vibrations are mostly affected by the area applying weight to the box's bottom. Thus, we define our model's target prediction to be the box's layout -- showing the approximate shape, orientation, and location of the inner object on the box floor (\cref{fig:setup}c).

Training a model to infer the contents of a general unknown box may require a large dataset of many boxes and internal object configurations. For the scope of this paper, we provide a proof of concept for a limited case: a model trained on three boxes and four different objects. Our main goal was to show that external box vibrations contain enough signal to localize the object layout within the box for new, unseen object positions, given the boxes and objects the model has seen. 
This scenario is useful since the set of possible containers and objects in many real-world applications may be finite (\eg, products in factory boxes). Nevertheless, we also show that our model can generalize beyond the training distribution and predict unseen object instances and orientations, multiple objects, and more.

\subsection{Capturing the data}
\label{sec:data_capture}
\figExperimentSetup
To collect data, we built an experimental rig consisting of an overhead camera, a vibration camera, a laser, and a set of speakers surrounding the box (see \cref{fig:setup}). We reproduced the laser and vibration camera as described in \cite{kichler2025learning} and refer the reader to the original work for further technical details. Without loss of generality, the vibration camera records $T$ vibration samples $\boldsymbol{v} \in \mathbb{R}^{100 \times 2 \times T}$ at sampling rate $f_{\rm samp}$ for a grid of $10{\times}10$ surface points on the side of the box facing the camera (\cref{fig:setup}d). At each of the $T$ frames, each of the $100$ laser points provides two measurements corresponding to the horizontal and vertical surface tilts of that laser speckle, up to a constant scale factor \cite{kichler2025learning}.\footnote{The constant that relates the measured speckle shifts in pixel units to degrees is approximately shared between all measured points.} For training, we leave the top of the box open so an overhead camera can capture the ground-truth box interior.
We use this overhead view only for supervision, so at inference time the box's interior is outside the line of sight and the box may be closed. 

To collect a data sample, we select a box and an object, and place the object at some position on the box floor. The overhead camera captures an RGB image of the box interior from a bird's-eye view (\cref{fig:setup}b). We crop the image to the box floor and segment the objects to obtain the ground-truth 2D occupancy layout $L {\in} \{0, 1\}^{H {\times} W}$, where $1$ indicates an area occupied by the object and $0$ is `free space' (\cref{fig:setup}c). Note that the layout $L$ may differ from the true physical object layout inside the box due to perspective distortion of the overhead camera. In this work, however, we neglect this effect and `expect' the model to learn to predict the distorted captured layout.

Different boxes have different shapes, which can yield different ground-truth layout sizes ${H \times W}$. However, we want our model to be agnostic to box shape. Therefore, we map each $L$ to one `standard domain':
\begin{equation}
    L^\text{s} \leftarrow {\cal M}(L),~~~ L^\text{s}{\in} \mathbb{R} ^{64 \times 64}
\end{equation}
where ${\mathcal{M}}$ is a bi-linear image resizing from ${H {\times} W}$ to a predefined and fixed size of a $64 {\times} 64$ image.

For each box and object configuration, we captured the box's vibrations (\cref{fig:setup}d) excited by different speaker positions in order to make our model invariant to the excitation source. Consequently, each data sample consists of a short vibration measurement (\eg, one second) and a ground-truth layout pair $(\boldsymbol{v}, L^\text{s})$. Given some predicted $\hat{L}^\text{s}$ layout, and the dimensions of the box captured using a conventional camera, we can apply $\mathcal{M}^{-1}$ to undo the image resizing and produce the predicted layout in the correct box dimensions.

\subsection{Learning the box's layout}
\label{sec:network}
We introduce WhisperBox, an encoder-decoder transformer that maps the vibration measurements $\boldsymbol{v}$ into a latent representation $z_{\text{latent}}$, which is decoded into a predicted 2D occupancy mask $\hat{L}^{\text{s}}$. The architecture consists of four main components: a tokenizer, a frequency encoder, a laser encoder, and a decoder (\cref{fig:network_arch}).

\figNetworkArch

\vspace{5pt}
\noindent\textbf{Tokenizer.}
We apply a preprocessing step $p$ to each vibration $\boldsymbol{v} \in \mathbb{R}^{100 \times 2 \times T}$ and then compute the one-sided discrete Fourier Transform (DFT) magnitude (per point and vibration axis)
\begin{equation}
 \boldsymbol{V} = |\boldsymbol H \odot\mathcal{F}(p(\boldsymbol{v}))|,~~~  \boldsymbol{V}{\in} \mathbb{R}^{100 \times 2 \times F},   
\end{equation}
where $\boldsymbol H{\in} \mathbb{C}^{1 \times 1 \times F}$ is some predefined high-pass filter and $F$ is the number of frequencies in a predetermined frequency range $[f_\text{min}, f_\text{max}]$. \footnote{We pad or subsample the temporal signals to fit the desired frequency range and bin size.} This formulation supports vibrations of arbitrary duration and sampling frequency, provided that $f_\text{samp}$ satisfies the Nyquist sampling rate and all frequencies lie within $[f_\text{min}, f_\text{max}]$. As in prior work, our model learns in the signal's spectral domain \cite{kichler2025learning, chen2021boombox}. 

We normalize each sample $\boldsymbol V$ to zero mean and unit variance over all laser points, vibration axes and frequency bins. Similar to image patches in Vision Transformers ~\cite{dosovitskiy2020image}, we split the $F$ frequency bins into $18$ non-overlapping bands of $65$ consecutive bins and interleave the two vibration axes within each band, giving a tensor of shape $(100, 18, 130)$. A shared linear project embeds each band into a $d$-dimensional token with total shape $(100, 18, d)$. 

\vspace{5pt}
\noindent\textbf{Frequency encoder.}
The frequency encoder learns relationships among the different vibration frequency bands. The frequency encoder is applied independently to each of the $100$ laser points, using the same transformer with 1D rotary positional embeddings (RoPE). For a single laser point $l$, the transformer models interactions across its 18 frequency bands, yielding features $h_{l,f}\in\mathbb{R}^d$. 

We aggregate these features into a single representation $z_l$ using a \textit{MergeBlock}, an operation that aggregates features in a token dependent way:
\begin{equation}
\label{eqn:frequency_merge_block}
    z_l
    =
    \sum_{f=1}^{18}
    W_f \mathcal{N}(h_{l,f}) + b_f
\end{equation}
where $\mathcal{N}$ denotes layer normalization and $W_f \in \mathbb{R}^{d\times d}, b_f \in \mathbb{R}^d$ are learned parameters. Unlike conventional pooling, which applies the same aggregation operation to every token, \textit{MergeBlock} uses a separate learned projection for each frequency band. This is particularly suitable for our setting because each token always corresponds to the same fixed frequency range. The model can therefore learn a distinct transformation for each band rather than treating frequency tokens interchangeably. We repeat this operation for all $100$ laser points and concatenate their representations, yielding $Z = [z_1,\ldots,z_{100}]^\top \in \mathbb{R}^{100\times d}$.

\vspace{5pt}
\noindent\textbf{Laser encoder.}
To recover the object's spatial layout, the laser encoder combines information across all the $100$ laser points in $Z$. A transformer uses 2D RoPE to encode the laser grid's spatial layout and produces a feature $h_l \in \mathbb{R}^d$ for each laser point $l$. A MergeBlock then aggregates these features into a global representation:
\begin{equation}
\label{eqn:laser_merge_block}
    z_\text{latent} =
    \sum_{l=1}^{100}
    W_l {\cal N}(h_l) + b_l
\end{equation}
where $W_l \in \mathbb{R}^{d \times e}, b_l \in \mathbb{R}^{e}$ are learned parameters. The resulting $z_{\text{latent}} \in \mathbb{R}^e$ is a global feature that captures the information about the hidden object across all $100$ laser points and each of their $18$ frequency bands. 

\vspace{5pt}
\noindent\textbf{Layout decoder.}
The decoder reconstructs the hidden object layout from the latent vector. First, a linear layer maps $z_\text{latent}$ to a coarse $4 \times 4$ feature map with $512$ channels. Four upsampling blocks then progressively double the spatial resolution and halve the number of channels, going from $4 \times 4 \times 512$ to $64 \times 64 \times 32$. Every block uses $2\times$ nearest-neighbor interpolation for upsampling, followed by two $3 \times 3$ convolutions each followed by group normalization and SiLU activations. A residual connection adds a $1 \times 1$ convolution of the upsampled input to the block output. After all the blocks, three $3 \times 3$ convolutions refine the features and produce a $64 \times 64$ occupancy layout map in $\mathbb{R}$. A final sigmoid converts these logits into probabilities, yielding the per-pixel occupancy layout $\hat{L}^s \in [0, 1]^{64 \times 64}$.

\vspace{5pt}
\noindent\textbf{Auxiliary classification heads.}
We also train two MLPs to predict the object class $\hat{o}$ and box class $\hat{b}$ from $z_\text{latent}$, providing additional supervision during training.

\vspace{5pt}
\noindent\textbf{Training objective.}
We train the network to predict the 2D occupancy layout using a combination of binary cross-entropy (BCE) and Dice loss \cite{milletari2016v}: 
\begin{align}
    \mathcal{L}_\text{layout}
    &=
    \mathcal{L}_\text{BCE}(\hat{L}^s, L^s)
    +
    \mathcal{L}_\text{Dice}(\hat{L}^s, L^s)
\end{align}
where the BCE loss encourages accurate pixel-wise predictions, while the Dice loss encourages shape overlap between the predicted and ground-truth layouts. The total loss is
\begin{align}
    \mathcal{L}_\text{total}
    &=
    \lambda_1 \mathcal{L}_\text{layout}(\hat{L}^s, L^s)
    +
    \lambda_2 \mathcal{L}_\text{CE}(\hat{b}, b)
    +
    \lambda_3 \mathcal{L}_\text{CE}(\hat{o}, o)
\end{align}
where $\mathcal{L}_\text{CE}$ is the cross-entropy loss and $\lambda_1, \lambda_2, \lambda_3 \in \mathbb{R}_+$ control the strength of the layout, box, and object losses. 

%% file: sec/4_details.tex
\section{Implementation details}
\label{sec:details}

\figQualitativeResults

\subsection{Experimental setup details}
For sensing vibrations, we replicated the speckle-based camera setup of Kichler \etal \cite{kichler2025learning}. The $10 \times 10$ laser grid was created using a HOLO-OR beamsplitter \cite{holoor_beam_splitters}. The boxes were placed approximately \qty{30}{\centi\metre} from the camera, so the projected laser grid covers as much of the box wall as possible without extending past the edges.

To capture the ground truth box layout, we installed an IDS UI-3070CP overhead camera at $\approx 100$ cm above the box. A rig of eight Creative Pebble V2 speakers surrounds the box. During data capture, we activated these speakers individually via an audio interface \cite{motu_ultralite_mk5}. We detached both the speakers and the camera from the optical table containing the probed box to prevent mechanical vibration transfer.

\subsection{Training details}
\label{sec:training}

\noindent\textbf{Ground-truth masks.}
For each sample, we use SAM-3 to obtain an object segmentation mask \cite{carion2026sam} as our ground truth label of our box layout. Since the bird's-eye view of our overhead camera is out-of-distribution for the segmentation model, we prompt primarily by color and shape rather than class. We use OpenCV's image upsampling and downsampling operations as $\mathcal{M}$ and $\mathcal{M}^{-1}$, respectively.

\vspace{5pt}
\noindent\textbf{Input signals.}
For each sample, we excite the box with a 1.3-second logarithmic chirp spanning $[f_\text{min}, f_\text{max}] = [100, 1000]$ Hz. We record the resulting vibrations at $f_\text{samp} = 2500$ Hz for $T=3250$ frames. The preprocessing function $p$ applies a zero-phase, fifth-order Butterworth band-pass filter over $[f_\text{min}, f_\text{max}]$, followed by a Tukey window ($\alpha=0.02$). We then compute the DFT at $F=1170$ predefined frequency bins $\{100.77, 101.54 ..., 1000.00\}$ Hz with increments of $0.77$ Hz. During training, we apply data augmentation as described in the supplementary.

\vspace{5pt}
\noindent\textbf{Model architecture.}
We set the frequency-band token dimension to $h=256$ and the latent dimension to $e=512$. The frequency and laser encoders contain four and eight transformer blocks, respectively. Each block uses eight attention heads, a GELU MLP with a 4× expansion ratio, and RoPE with a base of $100$. The auxiliary heads use a $512$-dimensional hidden layer with ReLU activation. We set the loss weights to $\lambda_1 = 0.5$ for the layout loss and $\lambda_2 = \lambda_3 = 0.1$ for the auxiliary losses. The resulting model has $34.5$ million parameters.

\vspace{5pt}
\noindent\textbf{Training.}
We train for $300$ epochs with AdamW (learning rate $10^{-4}$, weight decay $0.01$) using a 50-epoch warmup and cosine decay \cite{loshchilov2017decoupled}. With a batch size of 1024, training takes two hours on an RTX 5080.

%% file: sec/5_experimental.tex
\section{Experimental evaluation}
\label{sec:eval}

\subsection{Dataset}
We collected a dataset of 15,891 samples spanning three boxes, six objects, and five different speakers. For each object position, we collect four training samples corresponding to excitations from the same four speakers (one speaker active per sample) with an additional speaker reserved for testing.\footnote{The gastronome was recorded with three speakers, not four.} We selected three boxes commonly used in households and industry: a metal box, plastic box, and cardboard box (see \cref{fig:plots}). An X-ray machine cannot view the contents of a metal box because of its solid metal walls. The plastic box floor does not sit perfectly flat and curves up. The cardboard box has side flaps that vibrate with the audio, introducing noise. Our methods overcome all of these irregularities. For a more detailed overview of the dataset, see the supplementary material.

\begingroup
\setlength{\parskip}{4pt}

\subsection{Performance evaluation}
We evaluate the model's performance under two main generalization settings: in-distribution and out-of-distribution. The \textit{in-distribution} setting evaluates layout recovery for objects and object configurations seen during training, but placed at unseen positions within the box. This setting is relevant to industrial inspection applications where boxes and objects are standardized and belong to a finite set (\eg, the same product in packaging boxes). In the \textit{out-of-distribution} setting, we test performance across various scenarios not explicitly represented during training. We measure performance by Intersection over Union (IoU), which computes the overlap between the predicted and ground-truth object layout masks as a percentage.

\noindent\textbf{In-distribution evaluation.}
The in-distribution dataset consists of four single objects (candle, cube, soap dispenser, vase) with around $250$ positions in each box, and two multi-object scenes (two cubes, cube and vase).

\cref{fig:good_predictions}a shows that our model can accurately recover layouts for novel, unseen object positions. \cref{fig:good_predictions}b shows that the model can also predict multi-object layouts.
Nevertheless, our dataset contains significantly fewer two-object examples (around $100$ samples of cube-cube and cube-vase in each box), which correlates with the lower overall average IoU score for two-object prediction versus single-object prediction (see \cref{fig:plots}b). 

From vibration measurements alone, our model identifies objects and determines where they are inside the box. Object classification reaches $99\%$ accuracy for single-object scenes and $80\%$ for two-object scenes. Localization errors are $1.01 \pm 1.24$ cm and $3.45 \pm 4.76$ cm, respectively, computed as the L2 distance between predicted and ground-truth centers of mass extracted from the layout masks.

\noindent\textbf{Out-of-distribution evaluation.}
Here, we test our model, trained as described above, on a variety of cases that did not exist in the training set. \cref{fig:good_predictions}c shows layout predictions for unseen objects. These objects have a different weight, material composition, and exact geometry, yet our model can still localize their position and box layout fairly accurately (coffee pot and tomato sauce in \cref{fig:plots}b). \cref{fig:good_predictions}d shows that the model can handle unseen object rotations for non-symmetric objects. In this test, we excluded a range of rotation angles from the training set and applied inference for these cases. In \cref{fig:good_predictions}e, we use a similar strategy; however, here we exclude a domain of spatial box regions from the training and use it for testing.

In \cref{fig:good_predictions}f, we test inference on samples where the box was excited using the fifth speaker, not included during training. Much like Kichler \etal, this test shows that the model, trained on various acoustic excitations, can generalize to novel sound sources \cite{kichler2025learning}. Lastly, we test inference for a physically shifted laser grid location on the box. We tested two shift magnitudes: small (about 2cm) and large (about 9cm). \cref{fig:good_predictions}g shows the small-shift magnitude result, suggesting that our model can handle deviations in laser grid position as long as it roughly covers the same object surface region. However, large-scale shifts where the laser grid samples variations in parts of the boxes entirely unseen during training yield poor results (IoU $6\%$; see supp.).
\cref{fig:plots} summarizes all performance metrics across all tests.
\par
\endgroup

\figQuantitativeResults

%% file: sec/6_insights.tex
\section{Analysis and Insights}
\label{sec:insights}

\figInsights

\noindent\textbf{Frequency versus performance analysis.}
In \cref{fig:analysis}'s analysis, we ask: which frequencies contribute most to the predicted layout? And does the information lie near the resonant frequencies or elsewhere? 
To this end, for each box, we mask one frequency band token at a time ($65$ FFT bins, ${\approx}50$ Hz) and measure the resulting IoU drop, shown as bars in \cref{fig:analysis}. We also add the box's mean FFT magnitude.\footnote{Computed on $720$ in-distribution, single-object training samples: $60$ per (box, object) pair, $240$ per box.} 

The plots of \cref{fig:analysis} reveal two insights. First, in our model, no single frequency band holds the layout-related information. Instead, the model distributes its reliance across the spectrum: masking any one band costs at most $6\%$ of the IoU, and $15$--$16$ of the $18$ bands are needed for $90\%$ of the total sensitivity across all the boxes. 
Second, the bands whose removal most affects prediction \textit{do not} coincide with the object's resonant frequencies. This suggests that the model exploits layout-dependent spectral variations across a broad frequency range, rather than relying primarily on dominant box resonances.

\noindent\textbf{Latent space structure.}
In \cref{fig:analysis}b we take the $512$-dimensional latents of training samples from the plastic box under speaker $1$ and project them to 3D with UMAP. Each projection is drawn as a photo of its object and colored by the object's true position ($x$ is hue, $y$ is lightness).
The projected structure shows that although the model is trained only to predict segmentation masks and never explicitly sees coordinates, its latents encode position. This is evident because moving up and inward in the latent space moves up in the box $y$, while moving right and down moves right in the box $x$. 
The latents also cluster by object: a sample's nearest latent neighbor is the same object $85.6\%$ of the time (chance $17\%$, $6$ objects). 
Our dense sampling across the box floor yields a smooth embedding trajectory for each object, with color varying gradually with position.

\noindent\textbf{Latent space arithmetic.}
Word embeddings allow arithmetic such as \textit{king}${-}$\textit{man}${+}$\textit{woman}${=}$\textit{queen} \cite{mikolov2013linguistic}. 
\cref{fig:analysis}c shows our latent space has a similar property.
The top of \cref{fig:analysis}c  shows we can \textit{swap objects} by subtracting the mean latent of one object and adding the mean latent of another.
\cref{fig:analysis}c bottom shows that averaging two single-object latents and adding a vector that represents the interaction between a pair of objects yields the two objects' \textit{combined layout}.
See the supplementary for more details.

%% file: sec/7_discussion.tex
\section{Discussion and limitations}
\label{sec:discuss}

\vspace{5pt}
\noindent\textbf{Box and object assumptions.}
WhisperBox works well on metal, plastic, and cardboard boxes, but performs worse on wooden crates (see supplementary). We hypothesize that the wooden crates' thick walls had high damping, which suppresses the surface vibrations we measure. Moreover, wooden boxes are made by attaching several wooden plates, unlike the solid-piece containers that worked well, which may yield non-linear vibration responses. We used only objects that rest flush against the box floor. 
Objects with curved bottoms or objects cushioned in bubble wrap or packing material remain untested.

\vspace{5pt}
\noindent\textbf{Sensing assumptions.}
Our sensing relies on lasers, which can be unsafe for direct eye exposure, though this may be less of an issue in industrial settings. Specular or low-albedo materials, such as our metal box, are less amenable to speckle-based vibrometry. To boost the signal, we applied a small strip of white tape to the box wall to recover the speckle pattern.

\vspace{5pt}
\noindent\textbf{Closed boxes and dataset collection.}
Although trained only on open boxes to enable ground-truth imaging, our model \textit{did} recover layouts in some sealed boxes without any fine-tuning (see \cref{fig:analysis}d). Closing the lid changes the boundary conditions and vibrational response, degrading performance or causing failure, but since our method does not require an open lid, training on closed boxes could address this distribution shift. 
Naturally, our biggest limiting factor was the manual data collection. Automated data collection on assembly or packaging lines could overcome this bottleneck and help scale our method to more general predictions. 

%% file: sec/8_conclude.tex
\section{Conclusions}
\label{sec:conclude}

We introduce a method to reveal a box's contents by remotely sensing its vibrations and training a model called WhisperBox to predict its hidden interior layout. Unlike prior work, our method needs no physical contact to excite or sense vibrations. We experimentally show real-world layout predictions for various boxes and objects, as well as generalization to unseen objects and object configurations.

Future work may extend our method to more complex containers (\eg, a refrigerator) or densely packed, cluttered interiors (\eg, a fully packed suitcase). We envision applications in sensing objects on moving conveyor belts and scanning many boxes at once with a single laser grid. We hope our system, model, and dataset inspire new work on vibration-based perception and enable applications in security, industrial inspection, and warehouse shipping.